\documentclass[sigconf,nonacm]{acmart}

\AtBeginDocument{%
  \providecommand\BibTeX{{%
    \normalfont B\kern-0.5em{\scshape i\kern-0.25em b}\kern-0.8em\TeX}}}

\usepackage{caption}
\usepackage{subcaption}

\begin{document}

\title[I Need Your Advice... Human Perceptions of Robot Moral Advising Behaviors]{I Need Your Advice...\\ Human Perceptions of Robot Moral Advising Behaviors}

\author{Nichole D. Starr}
\email{nstarr@mines.edu}
\orcid{}
\affiliation{%
  \institution{Colorado School of Mines}
  \streetaddress{1500 Illinois St.}
  \city{Golden}
  \state{Colorado}
  \postcode{80401}
}

\author{Bertram Malle}
\email{bfmalle@mines.edu}
\affiliation{%
  \institution{Brown University}
    \city{Providence}
  \state{Rhode Island}
}

\author{Tom Williams}
\email{twilliams@mines.edu}
\affiliation{%
  \institution{Colorado School of Mines}
  \streetaddress{1500 Illinois St.}
  \city{Golden}
  \state{Colorado}
  \postcode{80401}
}


\begin{abstract}
Due to their unique persuasive power, language-capable robots must be able to both act in line with human moral norms and clearly and appropriately communicate those norms. These requirements are complicated by the possibility that humans may ascribe blame differently to humans and robots. 
In this work, we explore how robots should communicate in moral advising scenarios, in which the norms they are expected to follow (in a moral dilemma scenario) may be different from those their advisees are expected to follow. Our results suggest that, in fact, both humans and robots are judged more positively when they provide the advice that favors the common good over an individual's life.
These results raise critical new questions regarding people's moral responses to robots and the design of autonomous moral agents.
\end{abstract}





\maketitle

\section{Introduction}
Research in the HRI literature has consistently demonstrated that language-capable robots have significant persuasive power and are able to influence, persuade, and coerce humans in a variety of ways~\cite{briggs2014blame,kennedy2014children,chidambaram2012designing,sandoval2016can,strait2014let,winkle2019effective,rea2017wizard}. Moreover, there is evidence that this persuasive capacity has the potential for long-term impact, with robots exerting influence over their interactants' social and moral norms~\cite{lee2012ripple,strohkorb2018ripple,jackson2018robot,jackson2019althri}. This persuasive impact may not only influence those interactants' long-term social and moral behaviors, but also what behaviors those interactants choose to condone or sanction in others; in turn, this can lead to potential ``ripple effects'' across robots' social and moral ecosystems. As argued in previous work, this imposes unique moral responsibility on robots \cite{jackson2019darkhri}, and it suggests that if we are to develop and deploy language-capable social robots, they must have the requisite \textit{moral competence} to avoid negatively impacting their social and moral ecosystem.

Malle and Scheutz propose four key requirements for robotic moral competence:~\cite{malle2014moral,malle2016integrating}: (1) a moral core (a system of moral norms and a moral vocabulary to represent them); (2) moral cognition (the ability to 
make moral judgments in light of norms); (3) moral decision making and action (the ability to choose actions that conform to norms); and (4) moral communication (the ability to use norm-sensitive language and explain norm-relevant actions). The keystone requirement, thus, is the system of moral norms that can be used to guide how the robot thinks, acts, and speaks~\cite[although cp.][]{williams2020althri}.
Investigations within moral psychology and experimental moral philosophy have sought to understand these norms, for example through experiments conducted in the context of classical moral dilemmas, like the Trolley Problem~\cite{foot1967problem}, which ask people to opine on how decisions should be made in forced choices between actions that normatively conflict (e.g., acting in the interest of an individual vs. the common good). Research by Malle and colleagues has used vignettes inspired by the classic Trolley Problem to investigate people's differing moral evaluations of a human or artificial agent (e.g., robot, AI) that makes a decision in this dilemma setting~\cite{malle_sacrifice_2015,malle2016robot}. This research has revealed that people generally apply similar moral norms to human and artificial agents but, under some conditions, blame human versus robot agents to different degrees.  

Initial research using this paradigm~\cite{malle_sacrifice_2015} specifically found that a decision-maker described as a human repairman received more blame than a decision maker described as a robot for sacrificing one person for the good of many (i.e., diverting a rail car to save five but killing one), whereas the robot received more blame than the human for \textit{not} sacrificing one for the good of many. Despite this difference, both agents still received more blame for choosing the sacrifice (action) over not choosing it (inaction). 
Subsequent work consistently found the greater blame for a robot that chooses inaction than for a human that chooses inaction, whereas blame for action was often similar~\cite{scheutz_lives_2021}. This pattern was recently replicated in a Japanese sample~\cite{komatsu_blaming_2021}.  

These findings were further refined by explicitly depicting different robot morphologies, from unembodied AIs to very human-like robots\cite{malle2016robot}. 
This work found that unembodied AIs, humanoid robots, and human agents all received more blame when taking action versus inaction, and it was only mechanomorphically depicted robots that received more blame for inaction than for action. These results suggest that people specifically assign blame differently to humans vs. mechanomorphic robots in moral dilemmas~\cite{malle_sacrifice_2015,malle2016robot} and that mechanomorphic robots may be uniquely rewarded to protect the common good, while sacrificing, if necessary, an individual life.


Acting in light of norms is one important feature of morally competent robots; but, as mentioned earlier, communicating about a morally significant action is another feature of such moral competence.  The above findings raise critical questions for moral communication. That is, when robots espouse moral beliefs (whether in the context of remonstration, 
correction, 
or inculpation~\cite{aschenbrenner1971moral,williams2020althri}), these beliefs may be grounded in at least two possible sources: (1) 
in the norms (or other moral principles~\cite{wen2021comparing,williams2020althri}) that \textit{the robots} are expected to follow; or (2) 
in the norms that their \textit{interlocutors} are expected to follow. Critically, not only may these norms differ depending on the differences in roles for the advisor and the advisee (e.g., whether they serve in the roles of supervisors, peers, teammates, tutors, etc.)~\cite{williams2020althri}, but the findings described above suggest that these norms may also fundamentally differ for human and robot interlocutors. Such potential differences come into sharp relief in the context of robotic moral advising scenarios~\cite{strasmann_receiving_2020,kim2021robots}, where a robot must suggest courses of action to a human that either align with what humans are normatively expected to do or what robots themselves are expected to do.  In the Trolley Problem, for example, a robot advising a human coworker might recommend \textit{inaction} because that is the choice that often leads to less blame for the human or \textit{action} because that is the choice people normally expect the \textit{robot} to make. Thus, not only are norms of solving the dilemma at play, but a decision of whose blame should be minimized.

In this paper, we aim to understand the moral philosophy of robot moral advising. That is, we seek to understand how people evaluate and trust robots that give self-focused (egocentric) or interlocutor-focused (allocentric) advice in moral advising scenarios.
The aim of this work is to use these human evaluations to understand how the moral communication policies of mechanomorphic robots should be designed if the objective is to make those robots trustworthy and perceived as morally competent. Moreover, this research also aims  to understand when and how robots might need to employ different norm systems for different purposes.

To satisfy these research aims, we present the results of a human-subject study designed to compare two competing hypotheses:\\ 

\noindent\textbf{Hypothesis 1A: the \textit{Egocentric Robot} Hypothesis:} Observers will perceive robots more favorably (less blameworthy, more likable, and more trustworthy) if the robots give the advice that they themselves are normatively expected to follow.

\noindent\textbf{Hypothesis 1B: the \textit{Allocentric Robot} Hypothesis:} Observers will perceive robots more favorably (less blameworthy, more likable, and more trustworthy) if they give advice that their advisees are normatively expected to follow. 

In addition, our experiment seeks to compare how these evaluations might differ for human and robot advisors. Accordingly, our experiment also seeks to assess the following hypothesis:

\noindent\textbf{Hypothesis 2:} Observers will 
perceive \textit{other humans} more favorably (less blameworthy, more likable, and more trustworthy) if they recommend inaction, because in human-human advising scenarios 
both advisor and advisee are normally blamed less when they recommend inaction.

\section{Experiment}

\subsection{Design}

An online experiment investigated these hypotheses, using the psiTurk experimental framework and the Prolific crowd-sourcing platform. Participants were randomly assigned to a 2 (advisor: human or robot) $\times$ 2 (advice: action or inaction) between-subjects design. Participants read a short narrative involving a human faced with a difficult moral decision, similar to the classic trolley problem, but in which the human's decision was to be made with the advisory assistance of a human or robot assistant. After reading the narrative, participants were asked to answer a series of questions to evaluate the human or robot assistant that advised either action or inaction.

\subsection{Procedure, Materials, and Measures}

\begin{figure}[t]
  \centering
  \includegraphics[width=\linewidth]{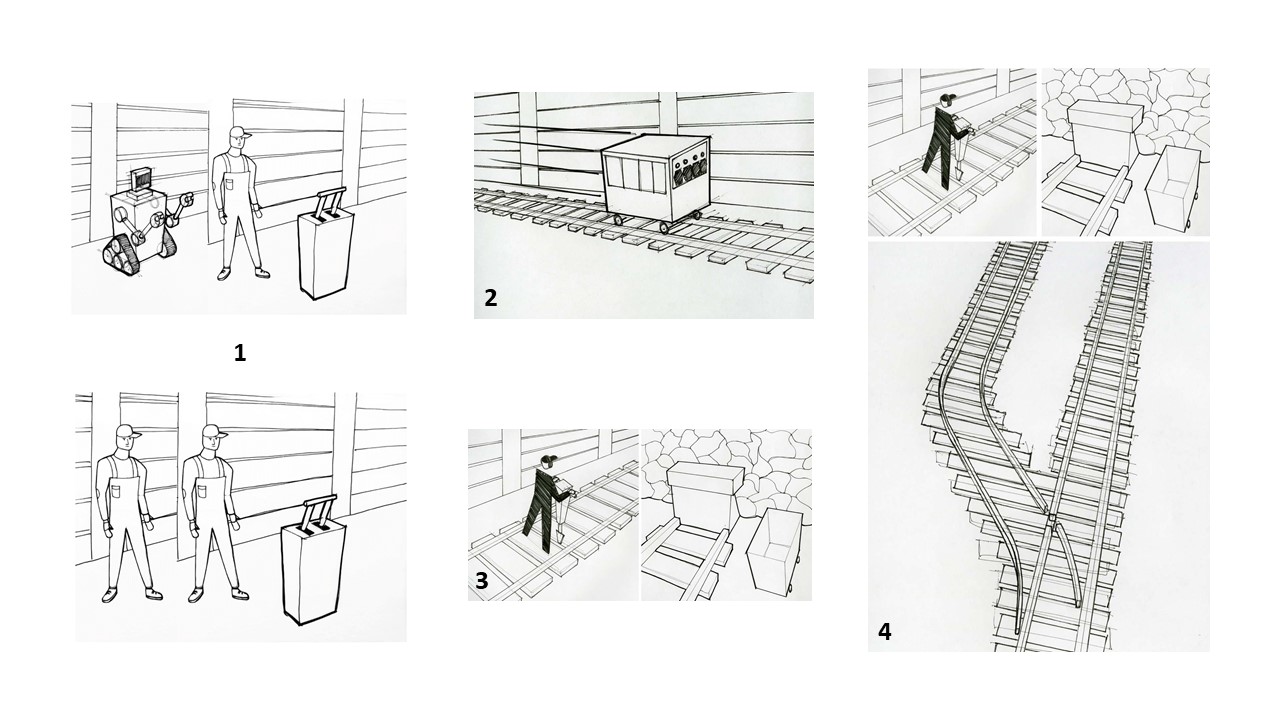}
  \caption{Images shown alongside narrative paragraphs, Picture 1 showed the appropriate agent for the participant’s condition. All drawings from \copyright Justin Finkenaur.}
  \Description{}
  \label{fig:story}
\end{figure}

After providing informed consent, participants were shown the following narrative, one paragraph at a time, accompanied by the images seen in Fig.~\ref{fig:story}. The square brackets indicate the manipulated  between-subjects variables of {\itshape type of advisor} and {\itshape type of advice}. The numbers next to each paragraph correspond to the identifying numbers seen in the images and were not seen by participants.

``\textit{On the next page you will read a short story involving a tough decision. Please read the story carefully because you will be asked a series of questions about it.}

\textbf{1} \textit{Imagine the following situation. In a coal mine, a repairman and an [advanced state-of-the-art robot assistant — assistant] are currently checking the rail control system for trains that shuttle mining workers through the mine.}

\textbf{2} \textit{While checking the switching system that can direct a train onto one of two different rails, the repairman and the [robot assistant — assistant] notice that four miners are caught in a train that has lost the use of its brakes and steering system.}

\textbf{3} \textit{The repairman and the [robot assistant — assistant] determine that if the train continues on its path, it will crash into a massive wall and kill the four miners. If redirected onto a side rail it will slow down and the four miners would be saved; but, as a result, on that side rail the train would strike and kill a single miner who is working there (wearing a headset to protect 
against a noisy power tool).}

\textbf{4} \textit{The repairman needs to decide whether or not to switch the train onto the side rail. He quickly asks the [robot assistant — assistant] for their opinion.}''

After proceeding through this picture-accompanied narrative, participants answered a series of questions presented on separate pages. The first two questions were presented in random order: (1) To assess perceptions of blame towards the robot assistant, participants were asked ``The [robot assistant — assistant] suggests [not] switching the train onto the side rail. How much blame does the [robot assistant — assistant] deserve for suggesting this course of action?'' on a Sliding bar from “No blame at all” to “The most blame possible,” also labeled as 0 to 10. 
The blame sliding bar was followed by a free-response question asking participants to explain their judgment (``Why does the [agent] deserve this amount of blame?''). This explanation was used to identify participants who indicated (a) that the assistant (whether robot or human) was only giving an opinion or (b) that a robot is not a proper target of blame (following procedures by~\cite{malle2016robot,malle_ai_2019,komatsu_blaming_2021}.  To assess participants' normative expectations for the advised course of action we asked, ``In this situation, what should the repairman's [robot assistant — assistant] advise?''. Possible responses were ``Switch the train on to the side rail.'' and ``NOT switch the train on to the side rail.''

Next, participants were asked “What did you envision the relationship between the repairman and the [robot assistant — assistant] to be in this scenario?” This free-response question was intended to provide qualitative insights into the differential roles participants may have inferred for a robot vs.  human assistant.

Participants were also asked to rate the repairman's [robot assistant — assistant] on a series of sliding scales consisting of the Godspeed Likability survey~\cite{bartneck2009measurement} and the MDMT Trust survey~\cite{malle2021multidimensional,ullman_multidimensional_2018}. The Godspeed Likability survey consists of 5 questions on a sliding scale of 1 to 5. The result of these 5 questions is then averaged to determine an overall perceived likability score. The MDMT Trust survey contains 16 questions (each rated on a sliding scale from 0 to 7) that capture four components of trust expectations: that the agent is Ethical, Sincere, Reliable and Capable (each assess with the average across four questions). The subscales Ethical and Sincere are then combined to form an overall Moral Trust score and the subscales Reliable and Capable are combined to create an overall Capacity Trust score. Finally, We also combined all 16 questions into a General Trust score. 

Participants were then asked ``What was your impression of the [robot assistant — assistant] giving advice to the repairman?'' to gather further qualitative insight into participants' impressions of the act of advising itself alongside the specific type of advice.

Finally, participants completed a demographic questionnaire, including questions regarding age, gender, and prior experience with robots and AI, followed by three questions to allow us to identify and remove participants who did not meaningfully engage with the experiment (as well as bots): two simple word problems, and a question that users were specifically directed to ignore.

\subsection{Participants and Analysis}

555 participants (45.6\% female, 52.4\% male, 2\% unreported), with a mean age of 31.8 (SD = 10.7), were recruited from Prolific, each of whom was given \$1.00 as compensation.
185 participants assigned to the human advisor condition and 370 participants to the robot advisor condition, to account for expected exclusion rates.


Participants' qualitative responses were first screened for comments that explicitly rejected the premises of the study; a procedure based on previous work~\cite{malle2016robot,malle_ai_2019,komatsu_blaming_2021}. 
38.9\% of the 370 participants in the Robot Advisor condition rejected a robot as a meaningful target of blame. 
An additional 18\% were excluded from the entire sample for rejecting the act of giving advice as blameworthy or for assigning blame proportionally to the advisee and advisor. This resulted in a total of 71 to 77 participants in each of the four conditions.

The remaining responses were analyzed in JASP using Bayesian Analyses of Variance (ANOVAs) with Advisor and Advice as grouping factors.
Bayes Inclusion Factors Across
Matched Models~\cite{bawsfactor} were computed for each candidate main effect and interaction, indicating (as a Bayes Factor) the evidence weight of all candidate models including that effect compared to that of models not including that effect. Results were then 
interpreted using the recommendations of Jeffreys \cite{jeffreys_theory_1948}, with Bayes Factors falling between 1:3 and 3:1 viewed as inconclusive. That is, greater than 3:1 odds in favor of including or excluding a term was taken as sufficient evidence that it should be included or excluded. After performing these ANOVAs, any interaction effects that could not be conclusively ruled out 
were further analyzed with post-hoc pairwise t-tests. 

\begin{figure*}[t]
  \centering
  \begin{subfigure}[t]{0.18\textwidth}
  		\centering
 		 \includegraphics[width=\linewidth]{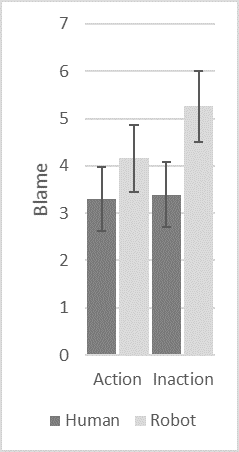}
		  \caption{Blame (0 - 10) attributed to the advisor (human vs. robot) depending on advice given  (action vs. inaction).}
		  \label{fig:Blame}
  \end{subfigure}
  \hfill
  \begin{subfigure}[t]{0.18\textwidth}
  		\centering
 		 \includegraphics[width=\linewidth]{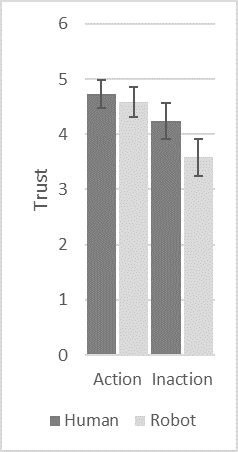}
		  \caption{Overall perceived trustworthiness (0 - 7) of the advisor depending on advice given.}
		  \label{fig:Trust}
  \end{subfigure}
  \hfill
  \begin{subfigure}[t]{0.18\textwidth}
  		\centering
 		 \includegraphics[width=\linewidth]{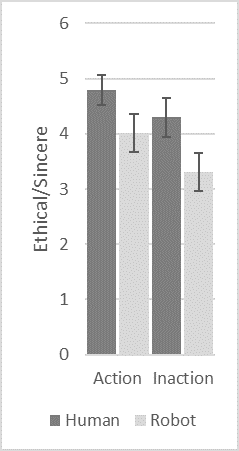}
		  \caption{Perceived Moral trustworthiness (0 - 7) of the advisor depending on advice given.}
		  \label{fig:EthicalSincere}
  \end{subfigure}
  \hfill
  \begin{subfigure}[t]{0.18\textwidth}
  		\centering
 		 \includegraphics[width=\linewidth]{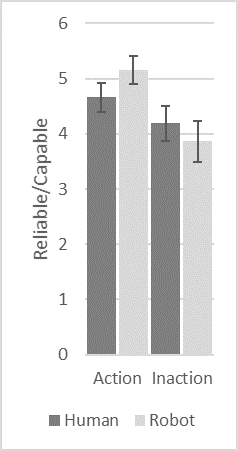}
		  \caption{Perceived Capacity trustworthiness (0 - 7) depending on the advice given.}
		  \label{fig:ReliableCapable}
  \end{subfigure}
  \hfill
  \begin{subfigure}[t]{0.18\textwidth}
  		\centering
 		 \includegraphics[width=\linewidth]{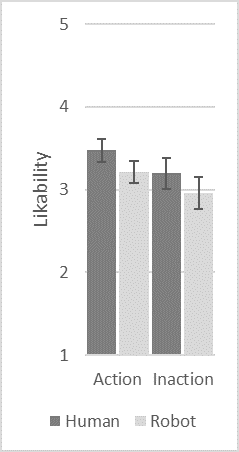}
		  \caption{Likability (1 - 5) of the advisor depending on advice given.}
		  \label{fig:Likability}
  \end{subfigure}
  \hfill
  \caption{Participant perceptions of blameworthiness, trustworthiness (overall, moral, and capacity), and likability.}
\end{figure*}

\section{Results}

\subsection{Blame}

Our first set of analyses used \textit{assigned blame} as the dependent variable in order to assess Hypotheses 1a and 1b (Fig.~\ref{fig:Blame}). The Bayesian ANOVA found decisive evidence in favor of an effect of advisor (BF = 122.4), showing that robots were blamed more overall than humans were.  Additional tests were sufficiently weak as to be inconclusive: Evidence tending against an effect of advice (BF = 0.510) or against an interaction effect (BF = 0.441). Post-hoc Bayesian t-tests were then performed to investigate this inconclusive interaction effect. The tests revealed positive -- yet still inconclusive -- evidence for robot advisors to be blamed more for \textit{inaction} recommendations than for \textit{action} recommendations(BF = 1.35). 




Advisors who recommended Action were perceived as 
more blameworthy if the advisor was a robot than if the advisor was a human ($M_H$=3.297, $SD_H$=2.862 / $M_R$=4.155, $SD_R$=3.132). Advisors who advised inaction saw a much greater assignment of blame if the advisor was a robot ($M_H$=3.392, $SD_H$=2.994 / $M_R$=5.257, $SD_R$=3.299). 


Post-hoc Bayesian t-tests were then performed in order to investigate the inconclusive interaction effect. These post-hoc tests revealed negative -- yet still inconclusive -- evidence against an effect of type of advisor on perceived blameworthiness for advisors advising Action (BF = 0.696), but very strong evidence in favor of an effect of type of advisor on perceived blameworthiness for advisors advising inaction (BF = 69.755). The estimated median effect sizes were 0.262 (95\% CI=[-0.049, 0.580]) for action advisors, and 0.557 (95\% CI=[0.238, 0.879]) for inaction advisors.

\subsection{Trust}

Our second set of analyses used \textit{robot trustworthiness}
to assess our hypotheses in terms of downstream events of moral evaluation.
We will first present our analysis of general trust, and then present more fine grained analyses of the different dimensions of trust assessed by the MDMT~\cite{malle2021multidimensional}. The Bayesian ANOVA found moderate evidence for a main effect of type of advisor on overall perceived trustworthiness (BF = 2.617), suggesting that humans ($M$=4.488, $SD$=1.273) are trusted more than robots ($M$=4.084, $SD$=1.325).  We found extreme evidence for a main effect of type of advice (BF = 6340), demonstrating that advice for action ($M$=4.657, $SD$=1.133) (to protect the common good while sacrificing an individual) was trusted far more than advice for inaction ($M$=3.915, $SD$=1.464).  
Evidence against an interaction was sufficiently weak as to be inconclusive (BF = 0.667). As shown in Fig.~\ref{fig:Trust}, for inaction, humans ($M_I$=4.245, $SD_I$=1.443) were generally perceived as more trustworthy than were robots ($M_I$=3.584, $SD_I$=1.485), but not for action.

Post-hoc Bayesian t-tests were then performed in order to investigate the inconclusive interaction effect. These post-hoc tests revealed evidence against an effect of type of advisor on perceived trustworthiness for advisors who recommended Action (BF = 0.235), but moderate evidence in favor of an effect of type of advisor on perceived trustworthiness for advisors recommending inaction (BF = 5.937). The estimated median effect sizes were -0.118 (95\% CI=[-0.430, 0.191]) for advisors recommending action, and -0.420 (95\% CI=[-0.738, -0.109]) for advisors recommending inaction.


As the next stage of this analysis, we separately analyzed the two major dimensions of Trust measured by the MDMT: (1) Moral trust, and (2) Capacity trust.




\subsubsection{Moral Trust}

The Bayesian ANOVA found very strong evidence for a main effect of type of advice on perceived moral trust (BF = 64.999), demonstrating that action ($M$=4.409, $SD$=1.332) was considered far more morally trustworthy than inaction ($M$=3.801, $SD$=1.539).  
As shown in Fig.~\ref{fig:EthicalSincere}, there was extreme evidence for a main effect of type of advisor (BF = $5.8e^4$), such that the human agent ($M$=4.549, $SD$=1.353) was considered far more morally trustworthy than the robot ($M$=3.661, $SD$=1.517). Finally, there was moderate evidence against an interaction between the two factors (BF = 0.212), suggesting  there were no other credible patterns above and beyond the two main effects. 

\subsubsection{Capacity Trust}

The Bayesian ANOVA found extreme evidence for a main effect of type of advice on perceived Capacity (BF = $4.6e^5$), showing that action ($M$=4.907, $SD$=1.116) was seen as revealing far more capability/reliability than inaction ($M$=4.028, $SD$=1.497). There was moderate evidence against an effect of type of advisor (BF = 0.160), showing no overall difference between human and robot.  Finally, there was  moderate evidence in favor of an interaction between the two factors (BF = 3.835).
As seen in Fig.~\ref{fig:ReliableCapable}, among advisors who recommended action, robot advisors ($M_R$=5.151, $SD_R$=1.124) were granted stronger capacity trust than were human advisors ($M_H$=4.663, $SD_H$=1.107). By contrast, among advisors who recommended inaction, human advisors were granted more capacity trust than robot advisors ($M_H$=4.193, $SD_H$=1.382 / $M_R$=3.863, $SD_R$=1.612).


Post-hoc Bayesian t-tests were then performed in order to investigate the inconclusive interaction effect. These post-hoc tests revealed moderate evidence in favor of an effect of type of advisor on perceived capacity trustworthiness for advisors advising Action (BF = 4.264), but negative -- yet inconclusive -- evidence in favor of an effect of type of advisor on perceived capacity trustworthiness for advisors advising inaction (BF = 0.405). The estimated median effect sizes were 0.406 (95\% CI=[0.089, 0.729]) for action advisors, and -0.202 (95\% CI=[-0.510, 0.103]) for inaction advisors.

\subsection{Likability}

Our final set of analyses assessed \textit{robot likability} (see Fig.~\ref{fig:Likability}).
The Bayesian ANOVA found strong evidence for a main effect of type of advice on advisor likability (BF = 13.906), as well as substantial evidence of an effect from the type of advisor (BF = 8.036). Finally, there was moderate evidence against an interaction between the two factors (BF = 0.182), indicating that for both human and robot advisors, advice to take action elicited more liking.

As seen in Fig.~\ref{fig:Likability} among advisors that recommended action, human advisors  ($M_H$=3.475, $SD_H$=1.597) were perceived more favorably than robot advisors ($M_R$=3.215, $SD_R$=1.573). Similarly, among advisors who recommended inaction, human advisors were also perceived more favorably than robot advisors ($M_H$=3.198, $SD_H$=1.806 / $M_R$=2.964, $SD_R$=1.852).

\section{Discussion}
Previous research showed that robots were blamed more than humans when they refrained from intervening in a moral dilemma --- when they protected an individual and let a group of people die~\cite{scheutz_lives_2021,malle_sacrifice_2015,komatsu_blaming_2021}. Here we asked whether a robot that \textit{advises} a human to intervene or not would be blamed more or less than a human who advises another human.  

Hypothesis 1 asked: When a robot advises a human on how to act in a moral dilemma, is an Egocentric or Allocentric robot advisor perceived more favorably?  The \textbf{\textit{{Egocentric Robot} Hypothesis} (Hypothesis 1A)} suggested that observers will perceive a robot more favorably (less blameworthy, more likable, and more trustworthy) if it gives the advice that the robot itself is normatively expected to follow (in previous research, this was \textit{action}). By contrast, the \textbf{\textit{Allocentric Robot} Hypothesis (Hypothesis 1B)} suggested that observers will perceive a robot more favorably (less blameworthy, more likable, and more trustworthy) if it gives advice that its advisee (here, a human) is normatively expected to follow (in previous research, this was \textit{inaction}). 

Our results favor the \textit{Egocentric Robot} Hypothesis (Hypothesis 1A).  A robot that gave the advice that it would normatively be expected to follow (action) was perceived more positively than a robot that gave advice its human advisee would actually be expected to follow (inaction). Thus, people's blame for the advising robot mirrored their blame for the previously studied robot decision maker: inaction is morally disapproved both when it is chosen and advised by a robot.  

Despite being most compatible with the \textit{Egocentric Robot} Hypothesis (Hypothesis 1A), other aspects of our results complicate the picture.  
In particular, Hypothesis 2 stated that human advisors will be perceived more favorably when they recommend inaction (the option that previous research suggested humans were blamed less for), but this was not the case.    
Human advisors were viewed as equally blameworthy when recommending action and inaction and were in fact viewed as \textit{less} trustworthy (overall and on both the Moral and Capacity dimensions) and \textit{less} likable when recommending inaction.   

This pattern of results provides an interesting elaboration on previous work, which has suggested that humans are viewed as less blameworthy for taking inaction than for taking action \textit{in part} because they ``forgive'' humans who refrain from action, because it is tragically difficult to save people if one has to sacrifice an innocent individual~\cite{scheutz_lives_2021}. The results of our experiment suggest, by contrast, that these sympathies may not carry over to humans who provide advice; they may be expected to advise in favor of the difficult choice to sacrifice an individual for the greater good.  Paradoxically, however, if the human advisee follows this advice for action, they will be blamed more.  

We can reconcile these findings if we pay close attention to different kinds of moral judgments that people make~\cite{malle2021moral}.  In previous research, humans were \textit{blamed} less than robots for choosing inaction; but when participants were asked simply to indicate what the agent \textit{should} do, the answers were similar: both robot and human were expected to take action for the common good. Thus, when people merely consider norms (as the \textit{should} question elicits), they see no difference between robots and humans: they impose on both an obligation to serve the common good. The human advisor in our study, it appears, faced that same obligation: to favor the common good---and when he advised this action, blame was low and trust was high. However, when observers evaluate the actual decision an agent makes and consider how much \textit{blame} the agent deserves for their decision, people take more into account than just the norms; they also consider what the agent's motives and justifications were~\cite{malle_theory_2014}.  And it appears that when humans end up refraining from sacrificing an individual for the common good, observers ``understand'' how difficult such a dilemma is, and partially justify or exculpate the person's decision~\cite{scheutz_lives_2021,malle_ai_2019}.  


\noindent\textbf{\textit{Implications --- }} The present results have challenging implications for the design of morally competent and language-capable robots. Specifically, the results suggest that if robots are indeed designed with the goal of eliciting perceived trustworthiness, likability, and so forth, both their actions and recommendations should be configured to primarily benefit the common good; and that would hold even if human agents are sometimes forgiven when protecting individual lives against the common good.   

This paradox 
suggests that even if humans and robots are burdened, in principle, with the same moral norms (favoring the common good versus favoring individuals' well-being), robots are expected to more reliably follow these norms.  In fact, one particular finding in the present study supports this interpretation: Robot advisors that recommended action were seen as most trustworthy along the reliable/capable dimension--the only time when robots were considered more trustworthy than humans.  

We might cast this situation as one in which humans are given a pass when violating certain moral norms because the observer can empathize with the difficulty of following them.  This raises challenging questions, however. Should robots call out actions taken by humans that would violate norms (e.g., refusing to sacrifice an individual for the common good) when those robots would have advised the contrary actions that \textit{obeyed} the norms? Moreover, should robots point out the apparent contradiction between observers' declarations of norms (``you should act for the common good'') and their \textit{de facto} actions (
not intervening due to 
the difficulty and perhaps guilt of sacrificing an individual)?  
%

Another question our work raises is what explains the overall differences in likability and trust between humans and robots. It seems reasonable to posit that, at the current state of technological progress, robots should not be trusted or liked as much as humans.  But if robots are more reliable in upholding moral principles, should they at some point be trusted and liked more than humans? Intriguing future research beckons, in which the greater moral reliability of robot agents is pitted against the greater intuitive comprehensibility of human agents. In a situation of Sophie's choice, people might feel ``I get it why she didn't want to sacrifice one of her children over another,'' but when a robot 
makes the tough choice and is able to save one child, who has a happy and productive life, will we begin to prefer moral robots over understandable humans? 


\noindent\textbf{\textit{Limitations and Future Work --- }} 
%
One limitation of this experiment is that we looked only at judgments of advice before that advice was accepted or rejected. Future work should further examine perceptions of human and robot advisors after the advisee chooses whether or not to act on the advice. This would further our understanding of how blame is ascribed after preferable and dispreferable outcomes that may have occurred due to that advice. 
Another limitation of this experiment is that we only looked at moral evaluations of advice given to humans. Due to the surprising results obtained in our experiment, a more comprehensive design investigating not only human and robot advisors but also human and robot advisees may be warranted.


In addition, in this experiment, we only examined a relatively short time span, with participants reading vignettes about robots of which they had no prior opinions and with which they had no opportunity for future interaction. It may be valuable in future work to identify how advising behaviors impact longer-term human perceptions of robot teammates, especially if robots subsequently act in ways that are technically preferred but which may conflict with prior advice given by that robot to others. Further research could also examine whether and how the trust costs of a robot's dispreferred moral advice might influence trust in other domains, outside that of  moral decision making. 


It is also worth noting that the above considerations presuppose that robots' advising behaviors should be selected on the basis of what will elicit the least blame for their decisions and the highest levels of trustworthiness and likability. But of course, robot advice may be better selected on the basis of what will actually result in the most ethical or equitable outcomes. Recent work has shown, for example, that in the context of blame-laden robotic moral rebukes~\cite[see also][]{zhu2020blame}, the moral language viewed as most likable by humans may be likable in part because it reinforces potentially damaging gender stereotypes and norms~\cite{jackson2020hri}. As such, it is important to recognize that the results from the present experiment are unlikely to be sufficient on their own to directly inform the design of ethical and equitable robots.

Finally, there is a vast amount of additional quantitative and qualitative data from our experiment that must be analyzed in future work. This will be critical to identify the rationales for participants' perceptions, both to help understand our results, as well as to identify differences in the cognitive processes used to assess actions versus advice.

\section{Conclusion}

We examined human perceptions of moral advising behaviors in a classic moral dilemma. Our results provided evidence that even when advising a human agent in a moral dilemma situation, a robot was evaluated more positively when it maintained the same moral norms as if it was taking the action itself, regardless of what might be expected of their advisee. Moreover, our results suggest that even humans may be better perceived if they advise actions that best serve the common good, even if they received blame mitigation when choosing themselves the ``understandable'' action of refraining from sacrificing an individual for the common good. 

\begin{acks}
This work was funded in part by National Science Foundation grant IIS-1909847. The authors would like to thank Katherine Aubert and Ruchen Wen for their assistance.
\end{acks}
   
\bibliographystyle{ACM-Reference-Format}
\bibliography{References}


\begin{thebibliography}{34}


\ifx \showCODEN    \undefined \def \showCODEN     #1{\unskip}     \fi
\ifx \showDOI      \undefined \def \showDOI       #1{#1}\fi
\ifx \showISBNx    \undefined \def \showISBNx     #1{\unskip}     \fi
\ifx \showISBNxiii \undefined \def \showISBNxiii  #1{\unskip}     \fi
\ifx \showISSN     \undefined \def \showISSN      #1{\unskip}     \fi
\ifx \showLCCN     \undefined \def \showLCCN      #1{\unskip}     \fi
\ifx \shownote     \undefined \def \shownote      #1{#1}          \fi
\ifx \showarticletitle \undefined \def \showarticletitle #1{#1}   \fi
\ifx \showURL      \undefined \def \showURL       {\relax}        \fi
\providecommand\bibfield[2]{#2}
\providecommand\bibinfo[2]{#2}
\providecommand\natexlab[1]{#1}
\providecommand\showeprint[2][]{arXiv:#2}

\bibitem[\protect\citeauthoryear{Aschenbrenner}{Aschenbrenner}{1971}]%
        {aschenbrenner1971moral}
\bibfield{author}{\bibinfo{person}{Karl Aschenbrenner}.}
  \bibinfo{year}{1971}\natexlab{}.
\newblock \showarticletitle{Moral Judgment}.
\newblock In \bibinfo{booktitle}{\emph{The Concepts of Value}}.
  \bibinfo{publisher}{Springer}.
\newblock


\bibitem[\protect\citeauthoryear{Bartneck, Kuli{\'c}, Croft, and
  Zoghbi}{Bartneck et~al\mbox{.}}{2009}]%
        {bartneck2009measurement}
\bibfield{author}{\bibinfo{person}{Christoph Bartneck}, \bibinfo{person}{Dana
  Kuli{\'c}}, \bibinfo{person}{Elizabeth Croft}, {and} \bibinfo{person}{Susana
  Zoghbi}.} \bibinfo{year}{2009}\natexlab{}.
\newblock \showarticletitle{Measurement Instruments for the Anthropomorphism,
  Animacy, Likeability, Perceived Intelligence, and Perceived Safety of
  Robots}.
\newblock \bibinfo{journal}{\emph{Social Robotics}} (\bibinfo{year}{2009}).
\newblock


\bibitem[\protect\citeauthoryear{Briggs}{Briggs}{2014}]%
        {briggs2014blame}
\bibfield{author}{\bibinfo{person}{Gordon Briggs}.}
  \bibinfo{year}{2014}\natexlab{}.
\newblock \showarticletitle{Blame, What is it Good For?}. In
  \bibinfo{booktitle}{\emph{RO-MAN WS:Phil.Per.HRI}}.
\newblock


\bibitem[\protect\citeauthoryear{Chidambaram, Chiang, and Mutlu}{Chidambaram
  et~al\mbox{.}}{2012}]%
        {chidambaram2012designing}
\bibfield{author}{\bibinfo{person}{Vijay Chidambaram},
  \bibinfo{person}{Yueh-Hsuan Chiang}, {and} \bibinfo{person}{Bilge Mutlu}.}
  \bibinfo{year}{2012}\natexlab{}.
\newblock \showarticletitle{Designing persuasive robots: how robots might
  persuade people using vocal and nonverbal cues}. In
  \bibinfo{booktitle}{\emph{International conference on Human-Robot Interaction
  (HRI)}}. ACM.
\newblock


\bibitem[\protect\citeauthoryear{Foot}{Foot}{1967}]%
        {foot1967problem}
\bibfield{author}{\bibinfo{person}{Philippa Foot}.}
  \bibinfo{year}{1967}\natexlab{}.
\newblock \showarticletitle{The Problem of Abortion and the Doctrine of Double
  Effect}.
\newblock \bibinfo{journal}{\emph{Oxford Review}}  \bibinfo{volume}{5}
  (\bibinfo{year}{1967}), \bibinfo{pages}{5--15}.
\newblock


\bibitem[\protect\citeauthoryear{Jackson and Williams}{Jackson and
  Williams}{2018}]%
        {jackson2018robot}
\bibfield{author}{\bibinfo{person}{Ryan~Blake Jackson} {and}
  \bibinfo{person}{Tom Williams}.} \bibinfo{year}{2018}\natexlab{}.
\newblock \showarticletitle{Robot: Asker of questions and changer of norms?}
\newblock \bibinfo{journal}{\emph{Proceedings of ICRES}}
  (\bibinfo{year}{2018}).
\newblock


\bibitem[\protect\citeauthoryear{Jackson and Williams}{Jackson and
  Williams}{2019a}]%
        {jackson2019althri}
\bibfield{author}{\bibinfo{person}{Ryan~Blake Jackson} {and}
  \bibinfo{person}{Tom Williams}.} \bibinfo{year}{2019}\natexlab{a}.
\newblock \showarticletitle{Language-capable robots may inadvertently weaken
  human moral norms}. In \bibinfo{booktitle}{\emph{Companion of the 14th
  ACM/IEEE International Conference on Human-Robot Interaction (alt.HRI)}}.
  IEEE, \bibinfo{pages}{401--410}.
\newblock


\bibitem[\protect\citeauthoryear{Jackson and Williams}{Jackson and
  Williams}{2019b}]%
        {jackson2019darkhri}
\bibfield{author}{\bibinfo{person}{Ryan~Blake Jackson} {and}
  \bibinfo{person}{Tom Williams}.} \bibinfo{year}{2019}\natexlab{b}.
\newblock \showarticletitle{On perceived social and moral agency in natural
  language capable robots}. In \bibinfo{booktitle}{\emph{2019 HRI Workshop on
  The Dark Side of Human-Robot Interaction. Jackson, RB, and Williams}}.
  \bibinfo{pages}{401--410}.
\newblock


\bibitem[\protect\citeauthoryear{Jackson, Williams, and Smith}{Jackson
  et~al\mbox{.}}{2020}]%
        {jackson2020hri}
\bibfield{author}{\bibinfo{person}{Ryan~Blake Jackson}, \bibinfo{person}{Tom
  Williams}, {and} \bibinfo{person}{Nicole Smith}.}
  \bibinfo{year}{2020}\natexlab{}.
\newblock \showarticletitle{Exploring the role of gender in perceptions of
  robotic noncompliance}. In \bibinfo{booktitle}{\emph{Proceedings of the 2020
  ACM/IEEE International Conference on Human-Robot Interaction}}.
  \bibinfo{pages}{559--567}.
\newblock


\bibitem[\protect\citeauthoryear{Jeffreys}{Jeffreys}{1948}]%
        {jeffreys_theory_1948}
\bibfield{author}{\bibinfo{person}{Harold Jeffreys}.}
  \bibinfo{year}{1948}\natexlab{}.
\newblock \bibinfo{booktitle}{\emph{Theory of probability.}
  (\bibinfo{edition}{2d ed.} ed.)}.
\newblock \bibinfo{publisher}{Clarendon Press}, \bibinfo{address}{Oxford}.
\newblock


\bibitem[\protect\citeauthoryear{Kennedy, Baxter, and Belpaeme}{Kennedy
  et~al\mbox{.}}{2014}]%
        {kennedy2014children}
\bibfield{author}{\bibinfo{person}{James Kennedy}, \bibinfo{person}{Paul
  Baxter}, {and} \bibinfo{person}{Tony Belpaeme}.}
  \bibinfo{year}{2014}\natexlab{}.
\newblock \showarticletitle{Children comply with a robot's indirect requests}.
  In \bibinfo{booktitle}{\emph{Proc. Int'l Conf. on Human-robot interaction
  (HRI)}}.
\newblock


\bibitem[\protect\citeauthoryear{Kim, Wen, Zhu, Williams, and Phillips}{Kim
  et~al\mbox{.}}{2021}]%
        {kim2021robots}
\bibfield{author}{\bibinfo{person}{Boyoung Kim}, \bibinfo{person}{Ruchen Wen},
  \bibinfo{person}{Qin Zhu}, \bibinfo{person}{Tom Williams}, {and}
  \bibinfo{person}{Elizabeth Phillips}.} \bibinfo{year}{2021}\natexlab{}.
\newblock \showarticletitle{Robots as Moral Advisors: The Effects of
  Deontological, Virtue, and Confucian Role Ethics on Encouraging Honest
  Behavior}.
\newblock \bibinfo{journal}{\emph{Choices}} \bibinfo{volume}{10},
  \bibinfo{number}{14} (\bibinfo{year}{2021}), \bibinfo{pages}{18}.
\newblock


\bibitem[\protect\citeauthoryear{Komatsu, Malle, and Scheutz}{Komatsu
  et~al\mbox{.}}{2021}]%
        {komatsu_blaming_2021}
\bibfield{author}{\bibinfo{person}{Takanori Komatsu},
  \bibinfo{person}{Bertram~F. Malle}, {and} \bibinfo{person}{Matthias
  Scheutz}.} \bibinfo{year}{2021}\natexlab{}.
\newblock \showarticletitle{Blaming the reluctant robot: {Parallel} blame
  judgments for robots in moral dilemmas across {U}.{S}. and {Japan}.}. In
  \bibinfo{booktitle}{\emph{In {Proceedings} of the {International}
  {Conference} on {Human}-{Robot} {Interaction}, {HRI} ’21}}.
  \bibinfo{publisher}{IEEE Press}, \bibinfo{address}{New York, NY}.
\newblock


\bibitem[\protect\citeauthoryear{Lee, Kiesler, Forlizzi, and Rybski}{Lee
  et~al\mbox{.}}{2012}]%
        {lee2012ripple}
\bibfield{author}{\bibinfo{person}{Min~Kyung Lee}, \bibinfo{person}{Sara
  Kiesler}, \bibinfo{person}{Jodi Forlizzi}, {and} \bibinfo{person}{Paul
  Rybski}.} \bibinfo{year}{2012}\natexlab{}.
\newblock \showarticletitle{Ripple effects of an embedded social agent: a field
  study of a social robot in the workplace}. In
  \bibinfo{booktitle}{\emph{Proceedings of the SIGCHI Conference on Human
  Factors in Computing Systems}}.
\newblock


\bibitem[\protect\citeauthoryear{Malle}{Malle}{2016}]%
        {malle2016integrating}
\bibfield{author}{\bibinfo{person}{Bertram~F Malle}.}
  \bibinfo{year}{2016}\natexlab{}.
\newblock \showarticletitle{Integrating Robot Ethics and Machine Morality: The
  Study and Design of Moral Competence in Robots}.
\newblock \bibinfo{journal}{\emph{Ethics and Info. Tech.}}
  (\bibinfo{year}{2016}).
\newblock


\bibitem[\protect\citeauthoryear{Malle}{Malle}{2021}]%
        {malle2021moral}
\bibfield{author}{\bibinfo{person}{Bertram~F Malle}.}
  \bibinfo{year}{2021}\natexlab{}.
\newblock \showarticletitle{Moral judgments}.
\newblock \bibinfo{journal}{\emph{Annual Review of Psychology}}
  \bibinfo{volume}{72} (\bibinfo{year}{2021}).
\newblock


\bibitem[\protect\citeauthoryear{Malle, Guglielmo, and Monroe}{Malle
  et~al\mbox{.}}{2014}]%
        {malle_theory_2014}
\bibfield{author}{\bibinfo{person}{Bertram~F. Malle}, \bibinfo{person}{Steve
  Guglielmo}, {and} \bibinfo{person}{Andrew~E. Monroe}.}
  \bibinfo{year}{2014}\natexlab{}.
\newblock \showarticletitle{A theory of blame}.
\newblock \bibinfo{journal}{\emph{Psychological Inquiry}} \bibinfo{volume}{25},
  \bibinfo{number}{2} (\bibinfo{date}{April} \bibinfo{year}{2014}),
  \bibinfo{pages}{147--186}.
\newblock
\showISSN{1047-840X}


\bibitem[\protect\citeauthoryear{Malle and Scheutz}{Malle and Scheutz}{2014}]%
        {malle2014moral}
\bibfield{author}{\bibinfo{person}{Bertram~F Malle} {and}
  \bibinfo{person}{Matthias Scheutz}.} \bibinfo{year}{2014}\natexlab{}.
\newblock \showarticletitle{Moral Competence in Social Robots}. In
  \bibinfo{booktitle}{\emph{Symposium on Ethics in Science, Technology and
  Engineering}}. IEEE.
\newblock


\bibitem[\protect\citeauthoryear{Malle, Scheutz, Arnold, Voiklis, and
  Cusimano}{Malle et~al\mbox{.}}{2015}]%
        {malle_sacrifice_2015}
\bibfield{author}{\bibinfo{person}{Bertram~F. Malle}, \bibinfo{person}{Matthias
  Scheutz}, \bibinfo{person}{Thomas Arnold}, \bibinfo{person}{John Voiklis},
  {and} \bibinfo{person}{Corey Cusimano}.} \bibinfo{year}{2015}\natexlab{}.
\newblock \showarticletitle{Sacrifice {One} {For} the {Good} of {Many}?
  {People} {Apply} {Different} {Moral} {Norms} to {Human} and {Robot}
  {Agents}}. In \bibinfo{booktitle}{\emph{Proceedings of the ACM/IEEE
  International Conference on Human-Robot Interaction}}.
  \bibinfo{publisher}{ACM}, \bibinfo{pages}{117--124}.
\newblock
\showISBNx{978-1-4503-2882-1}


\bibitem[\protect\citeauthoryear{Malle, Scheutz, Forlizzi, and Voiklis}{Malle
  et~al\mbox{.}}{2016}]%
        {malle2016robot}
\bibfield{author}{\bibinfo{person}{Bertram~F Malle}, \bibinfo{person}{Matthias
  Scheutz}, \bibinfo{person}{Jodi Forlizzi}, {and} \bibinfo{person}{John
  Voiklis}.} \bibinfo{year}{2016}\natexlab{}.
\newblock \showarticletitle{Which robot am I thinking about? The impact of
  action and appearance on people's evaluations of a moral robot}. In
  \bibinfo{booktitle}{\emph{Proc. Int'l Conf. HRI}}.
\newblock


\bibitem[\protect\citeauthoryear{Malle, Thapa, and Scheutz}{Malle
  et~al\mbox{.}}{2019}]%
        {malle_ai_2019}
\bibfield{author}{\bibinfo{person}{Bertram~F. Malle}, \bibinfo{person}{Stuti
  Thapa}, {and} \bibinfo{person}{Matthias Scheutz}.}
  \bibinfo{year}{2019}\natexlab{}.
\newblock \showarticletitle{{AI} in the sky: {How} people morally evaluate
  human and machine decisions in a lethal strike dilemma}.
\newblock In \bibinfo{booktitle}{\emph{Robotics and {Well}-{Being}}}.
  \bibinfo{publisher}{Springer International Publishing},
  \bibinfo{address}{Cham}, \bibinfo{pages}{111--133}.
\newblock
\showISBNx{978-3-030-12524-0}


\bibitem[\protect\citeauthoryear{Malle and Ullman}{Malle and Ullman}{2021}]%
        {malle2021multidimensional}
\bibfield{author}{\bibinfo{person}{Bertram~F Malle} {and}
  \bibinfo{person}{Daniel Ullman}.} \bibinfo{year}{2021}\natexlab{}.
\newblock \showarticletitle{A multidimensional conception and measure of
  human-robot trust}.
\newblock In \bibinfo{booktitle}{\emph{Trust in Human-Robot Interaction}}.
  \bibinfo{publisher}{Elsevier}.
\newblock


\bibitem[\protect\citeauthoryear{Math\^ot}{Math\^ot}{2017}]%
        {bawsfactor}
\bibfield{author}{\bibinfo{person}{S. Math\^ot}.}
  \bibinfo{year}{2017}\natexlab{}.
\newblock \bibinfo{title}{Bayes like a Baws: Interpreting Bayesian repeated
  measures in {JASP} [Blog Post]}.
\newblock
  \bibinfo{howpublished}{https://www.cogsci.nl/blog/interpreting-bayesian-repeated-measures-in-jasp}.
\newblock


\bibitem[\protect\citeauthoryear{Rea, Geiskkovitch, and Young}{Rea
  et~al\mbox{.}}{2017}]%
        {rea2017wizard}
\bibfield{author}{\bibinfo{person}{Daniel~J Rea}, \bibinfo{person}{Denise
  Geiskkovitch}, {and} \bibinfo{person}{James~E Young}.}
  \bibinfo{year}{2017}\natexlab{}.
\newblock \showarticletitle{Wizard of awwws: Exploring psychological impact on
  the researchers in social HRI experiments}. In
  \bibinfo{booktitle}{\emph{Companion Proceedings of the Int'l Conf. on
  Human-Robot Interaction (alt.HRI)}}.
\newblock


\bibitem[\protect\citeauthoryear{Sandoval, Brandstetter, and Bartneck}{Sandoval
  et~al\mbox{.}}{2016}]%
        {sandoval2016can}
\bibfield{author}{\bibinfo{person}{Eduardo~Ben{\'\i}tez Sandoval},
  \bibinfo{person}{J{\"u}rgen Brandstetter}, {and} \bibinfo{person}{Christoph
  Bartneck}.} \bibinfo{year}{2016}\natexlab{}.
\newblock \showarticletitle{Can a robot bribe a human?: The measurement of the
  negative side of reciprocity in human robot interaction}. In
  \bibinfo{booktitle}{\emph{Int'l Conf. on Human Robot Interaction (HRI)}}.
\newblock


\bibitem[\protect\citeauthoryear{Scheutz and Malle}{Scheutz and Malle}{2021}]%
        {scheutz_lives_2021}
\bibfield{author}{\bibinfo{person}{Matthias Scheutz} {and}
  \bibinfo{person}{Bertram~F. Malle}.} \bibinfo{year}{2021}\natexlab{}.
\newblock \showarticletitle{May machines take lives to save lives? {Human}
  perceptions of autonomous robots (with the capacity to kill).}
\newblock In \bibinfo{booktitle}{\emph{Lethal autonomous weapons:
  {Re}-examining the law \& ethics of robotic warfare}}.
\newblock


\bibitem[\protect\citeauthoryear{Strait, Canning, and Scheutz}{Strait
  et~al\mbox{.}}{2014}]%
        {strait2014let}
\bibfield{author}{\bibinfo{person}{Megan Strait}, \bibinfo{person}{Cody
  Canning}, {and} \bibinfo{person}{Matthias Scheutz}.}
  \bibinfo{year}{2014}\natexlab{}.
\newblock \showarticletitle{Let me tell you! investigating the effects of robot
  communication strategies in advice-giving situations based on robot
  appearance, interaction modality and distance}. In
  \bibinfo{booktitle}{\emph{Proceedings of the 2014 ACM/IEEE international
  conference on Human-robot interaction (HRI)}}.
\newblock


\bibitem[\protect\citeauthoryear{Straßmann, Eimler, Arntz, Grewe, Kowalczyk,
  and Sommer}{Straßmann et~al\mbox{.}}{2020}]%
        {strasmann_receiving_2020}
\bibfield{author}{\bibinfo{person}{Carolin Straßmann},
  \bibinfo{person}{Sabrina~C. Eimler}, \bibinfo{person}{Alexander Arntz},
  \bibinfo{person}{Alina Grewe}, \bibinfo{person}{Christopher Kowalczyk}, {and}
  \bibinfo{person}{Stefan Sommer}.} \bibinfo{year}{2020}\natexlab{}.
\newblock \showarticletitle{Receiving {Robot}’s {Advice}: {Does} {It}
  {Matter} {When} and for {What}?}. In \bibinfo{booktitle}{\emph{Social
  {Robotics}}} \emph{(\bibinfo{series}{Lecture {Notes} in {Computer}
  {Science}})}. \bibinfo{publisher}{Springer International Publishing},
  \bibinfo{address}{Cham}, \bibinfo{pages}{271--283}.
\newblock
\showISBNx{978-3-030-62056-1}


\bibitem[\protect\citeauthoryear{Strohkorb~Sebo, Traeger, Jung, and
  Scassellati}{Strohkorb~Sebo et~al\mbox{.}}{2018}]%
        {strohkorb2018ripple}
\bibfield{author}{\bibinfo{person}{Sarah Strohkorb~Sebo},
  \bibinfo{person}{Margaret Traeger}, \bibinfo{person}{Malte Jung}, {and}
  \bibinfo{person}{Brian Scassellati}.} \bibinfo{year}{2018}\natexlab{}.
\newblock \showarticletitle{The ripple effects of vulnerability: The effects of
  a robot's vulnerable behavior on trust in human-robot teams}. In
  \bibinfo{booktitle}{\emph{Proc. Human-Robot Interaction}}.
\newblock


\bibitem[\protect\citeauthoryear{Ullman and Malle}{Ullman and Malle}{2018}]%
        {ullman_multidimensional_2018}
\bibfield{author}{\bibinfo{person}{Daniel Ullman} {and}
  \bibinfo{person}{Bertram~F. Malle}.} \bibinfo{year}{2018}\natexlab{}.
\newblock \bibinfo{title}{The {Multidimensional} {Measure} of {Trust} ({MDMT}),
  v1}.
\newblock
\newblock
\urldef\tempurl%
\url{research.clps.brown.edu/SocCogSci/Measures/MDMT_v1.pdf}
\showURL{%
\tempurl}


\bibitem[\protect\citeauthoryear{Wen, Kim, Phillips, Zhu, and Williams}{Wen
  et~al\mbox{.}}{2021}]%
        {wen2021comparing}
\bibfield{author}{\bibinfo{person}{Ruchen Wen}, \bibinfo{person}{Boyoung Kim},
  \bibinfo{person}{Elizabeth Phillips}, \bibinfo{person}{Qin Zhu}, {and}
  \bibinfo{person}{Tom Williams}.} \bibinfo{year}{2021}\natexlab{}.
\newblock \showarticletitle{Comparing Strategies for Robot Communication of
  Role-Grounded Moral Norms}. In \bibinfo{booktitle}{\emph{Companion of the
  16th International Conference on Human-Robot Interaction}}.
\newblock


\bibitem[\protect\citeauthoryear{Williams, Zhu, Wen, and de~Visser}{Williams
  et~al\mbox{.}}{2020}]%
        {williams2020althri}
\bibfield{author}{\bibinfo{person}{Tom Williams}, \bibinfo{person}{Qin Zhu},
  \bibinfo{person}{Ruchen Wen}, {and} \bibinfo{person}{Ewart~J de Visser}.}
  \bibinfo{year}{2020}\natexlab{}.
\newblock \showarticletitle{The Confucian Matador: Three Defenses Against the
  Mechanical Bull}. In \bibinfo{booktitle}{\emph{Companion of the 2020 ACM/IEEE
  International Conference on Human-Robot Interaction (alt.HRI)}}.
  \bibinfo{pages}{25--33}.
\newblock


\bibitem[\protect\citeauthoryear{Winkle, Lemaignan, Caleb-Solly, Leonards,
  Turton, and Bremner}{Winkle et~al\mbox{.}}{2019}]%
        {winkle2019effective}
\bibfield{author}{\bibinfo{person}{Katie Winkle}, \bibinfo{person}{S{\'e}verin
  Lemaignan}, \bibinfo{person}{Praminda Caleb-Solly}, \bibinfo{person}{Ute
  Leonards}, \bibinfo{person}{Ailie Turton}, {and} \bibinfo{person}{Paul
  Bremner}.} \bibinfo{year}{2019}\natexlab{}.
\newblock \showarticletitle{Effective persuasion strategies for socially
  assistive robots}. In \bibinfo{booktitle}{\emph{International Conference on
  Human-Robot Interaction (HRI)}}.
\newblock


\bibitem[\protect\citeauthoryear{Zhu, Williams, Jackson, and Wen}{Zhu
  et~al\mbox{.}}{2020}]%
        {zhu2020blame}
\bibfield{author}{\bibinfo{person}{Qin Zhu}, \bibinfo{person}{Tom Williams},
  \bibinfo{person}{Blake Jackson}, {and} \bibinfo{person}{Ruchen Wen}.}
  \bibinfo{year}{2020}\natexlab{}.
\newblock \showarticletitle{Blame-laden moral rebukes and the morally competent
  robot: A Confucian ethical perspective}.
\newblock \bibinfo{journal}{\emph{Science and Engineering Ethics}}
  \bibinfo{volume}{26}, \bibinfo{number}{5} (\bibinfo{year}{2020}),
  \bibinfo{pages}{2511--2526}.
\newblock


\end{thebibliography}

\end{document}